\newcommand{\ignore}[1]{}

\documentclass[conference]{IEEEtran}
\IEEEoverridecommandlockouts
\usepackage{cite}
\usepackage{amsmath,amssymb,amsfonts}
\usepackage{algorithmic}
\usepackage{graphicx}
\usepackage{textcomp}
\usepackage{xcolor}
\usepackage{caption}
\usepackage{subcaption}
\def\BibTeX{{\rm B\kern-.05em{\sc i\kern-.025em b}\kern-.08em
    T\kern-.1667em\lower.7ex\hbox{E}\kern-.125emX}}
\begin{document}

\title{Improving Spiking Neural Network Accuracy\\ Using Time-based Neurons
\thanks{This research was supported by National R\&D Program through the National Research Foundation of Korea (NRF) funded by Ministry of Science and ICT(2021M3F3A2A01037928).}
}


\author{
        Hanseok Kim\IEEEauthorrefmark{1}\IEEEauthorrefmark{2} and Woo-Seok Choi\IEEEauthorrefmark{1}\\
            \IEEEauthorrefmark{1}\textit{Department of ECE, ISRC, Seoul National University,} Seoul, South Korea \\
            \IEEEauthorrefmark{2}\textit{Samsung Electronics,} Hawseong, South Korea  \\ 
            Email:\{anjeo, wooseokchoi\}@snu.ac.kr
}

\maketitle

\begin{abstract}
Due to the fundamental limit to reducing power consumption of running deep learning models on von-Neumann architecture, 
research on neuromorphic computing systems based on low-power spiking neural networks using analog neurons 
is in the spotlight. 
In order to integrate a large number of neurons, 
neurons need to be designed to occupy a small area,
but as technology scales down, analog neurons are difficult to scale,
and they suffer from reduced voltage headroom/dynamic range and circuit nonlinearities.
In light of this, this paper first models the nonlinear behavior 
of existing current-mirror-based voltage-domain neurons designed in a 28\,nm process, 
and show SNN inference accuracy can be degraded by the effect of neuron's nonlinearity. 
Then, to mitigate this problem, we propose a novel neuron,
which processes incoming spikes in the time domain and greatly improves the linearity,
thereby improving the inference accuracy compared to the existing voltage-domain neuron. 
Tested on the MNIST dataset, 
the inference error rate of the proposed neuron differs by less than 0.1\,\%
from that of the ideal neuron.
\end{abstract}

\section{Introduction}
\label{sec:intro}
Deep neural networks (DNNs), or artificial neural networks (ANNs), have evolved into a state-of-the-art approach for machine learning tasks, 
and increasing number of services, e.g. classification, searching, translation, and recommendation.
Exploiting ANNs in a growing number of applications demands techniques for efficient implementation of ANN algorithms.
However, realizing ANN algorithms on existing computing systems poses formidable challenges.
Specifically, in conventional von-Neumann architectures,
memory access dominates the latency and energy cost for realizing ANN algorithms~\cite{horowitz20141}.

As an alternative to the von-Neumann architecture, 
neuromorphic systems based on spiking neural networks (SNNs) imitating the human brain 
are attracting attention for low-power deep learning hardware. 
In particular, architectures that employ non-volatile memory crossbars and SNNs 
have gained interests due to the fact that
1) no memory access is required, and 2) implementing matrix-vector multiplication, which is a ubiquitous operator in machine learning applications, is simple.
Each neuron of the SNN can output a spike, and the frequency of occurrence of this spike can be considered the real-valued neuron output of the ANN. 
This provides the SNN-based system with another feature, an event-driven system: 
it operates only when an event occurs, i.e. system stays in an idle state when no event occurs, which leads to power reduction in many practical scenarios. 

Unfortunately, however, in spite of these favorable aspects in hardware, SNNs have not been widely used as ANNs 
because SNNs are difficult to train, 
and the performance of SNNs has been worse than that of ANNs.
Since SNNs exploit binary activation functions to generate spikes 
and the derivative of the binary activation is zero almost everywhere,
conventional gradient-based learning cannot be applied.
Various workaround methods such as \cite{neftci2019surrogate,mostafa2017supervised,shrestha2018slayer,wu2018spatio, Ref:4}
have been proposed to train high-performance SNNs,
but their performance has not been scaled as well as ANNs for deeper and larger models.
Recently works such as \cite{Ref:3,hwang2020impact} 
showed that, instead of training SNNs from scratch, 
a high-performance SNN can be obtained by converting a trained ANN into an equivalent SNN using integrate-and-fire (IF) neurons.

\begin{figure}
\begin{subfigure}[b]{.5\textwidth}
    \centering
    \includegraphics[scale=0.45]{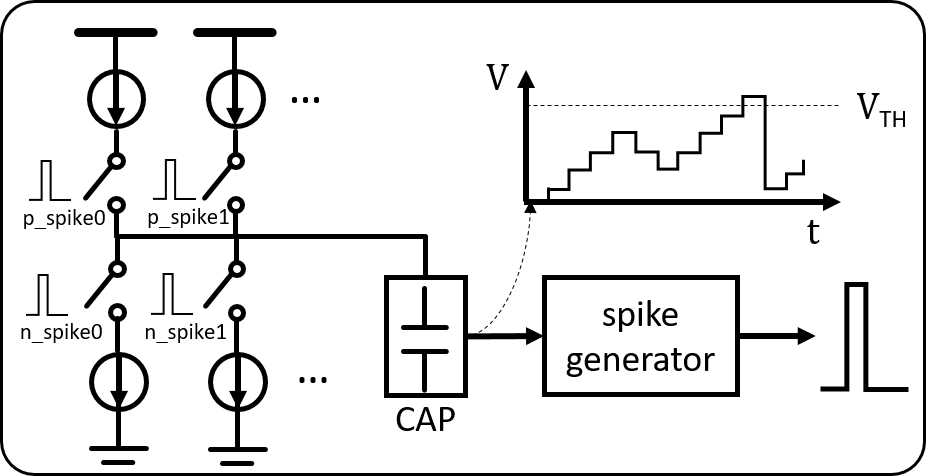}
    \caption{Conventional current-mirror-based voltage-domain neuron}
    \label{fig:vd}
\end{subfigure}
\hfill

\begin{subfigure}[b]{0.5\textwidth}
    \centering
    \includegraphics[scale=0.45]{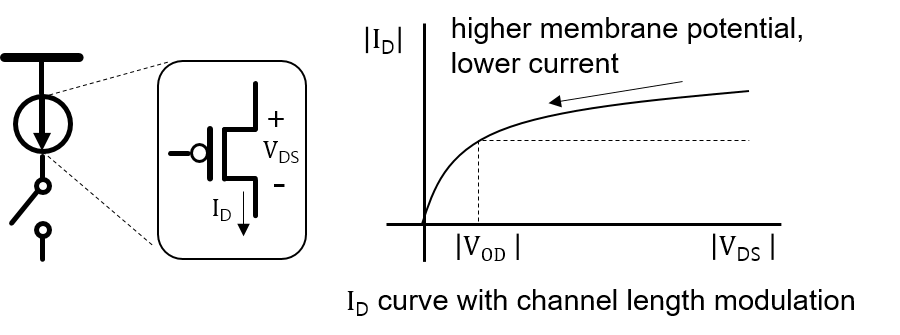}
    \caption{Nonlinearity of voltage-domain neuron}
    \label{fig:vd_nonideal}
\end{subfigure}
\hfill

\begin{subfigure}[b]{0.5\textwidth}
    \centering
    \includegraphics[scale=0.45]{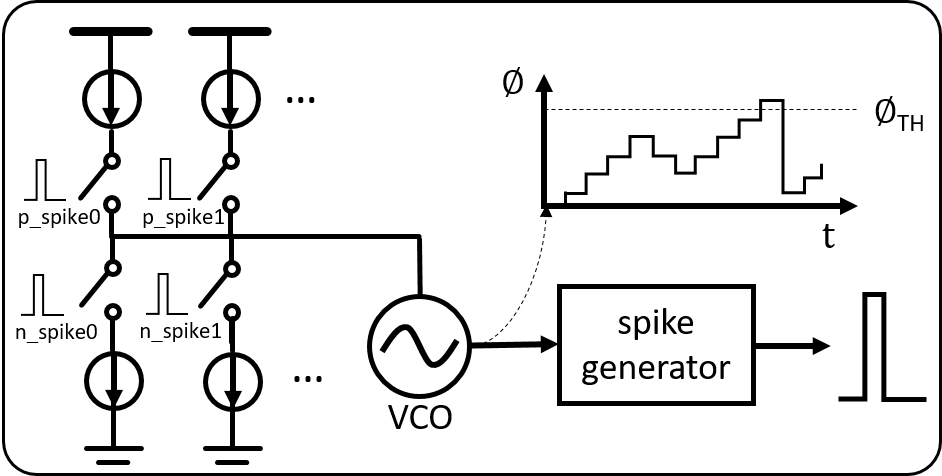}
    \caption{Proposed time-domain neuron}
    \label{fig:td}
\end{subfigure}
\hfill
\caption{Conceptual diagrams for conventional and proposed neurons.}
\label{fig:vdtd}
\end{figure}

Such integrate-and-fire neurons can be implemented using either digital circuits or analog circuits. 
In the case of digital neurons, the desired computation such as multiply and accumulate (MAC) can be accurately performed without any error, whereas it requires many transistors consuming power and area. 
Moreover, synchronous digital systems need power-hungry global clock distribution. 
On the other hand, when neurons are implemented using analog circuits, 
MAC can be implemented with relatively fewer transistors. 
For instance, analog SNN neurons proposed in \cite{Ref:1,hwang2020impact} (depicted in Fig.~\ref{fig:vdtd}(a)) are voltage-domain neurons, i.e. membrane potential is mapped to voltage, 
where spike inputs are converted into current and accumulated in a capacitor. 
Although simple, conventional current-mirror-based voltage-domain neurons suffer from  nonlinear behavior due to the channel length modulation of the transistor in deep submicron technologies.
 
In this paper, we first investigate the nonlinearity of the voltage-domain analog neurons in a 28\,nm CMOS process, 
and model its behavior to see how much it degrades the performance, or SNN inference accuracy. 
In order to mitigate the nonlinear problem, we propose a new time-based analog neuron,
which is also modeled and simulated at a system level to compare its performance with the existing scheme. 
Simulation results demonstrate that analog neuron's nonlinear behavior can degrade the performance even for small SNNs 
and the proposed SNN using time-based neurons shows negligible accuracy difference from that using ideal neurons.  

\section{Background}
\label{sec:background}

\subsection{ANN-to-SNN conversion}
It is known that training SNN with traditional backpropagation is difficult~\cite{huh2017gradient}. 
In spite of various studies on training SNN\cite{Ref:4,Ref:5}, the performance has not yet reached the ANN. 
Instead, many studies have been conducted to convert trained ANN parameters to SNN parameters,
so the SNN can have the same performance as the trained ANN~\cite{Ref:6,Ref:7}. 
In \cite{Ref:3}, through effective parameter normalization, the accuracy of the converted SNN approaches that of the original ANN. 
Despite these achievements, 
it is challenging to maintain accuracy in practice
since analog circuits suffer from noise, reduced voltage headroom/dynamic range, and nonlinearity.
Among many potential nonideal factors, we will focus on nonlinearity and discuss how it manifests in ANN-to-SNN conversion.

\subsection{Nonlinearity of voltage-domain neuron}

The conceptual diagram of the conventional current-mirror-based voltage-domain neuron~\cite{Ref:1,hwang2020impact} is described in Fig.~\ref{fig:vdtd}(a).
When a spike occurs, a current flows into or out of the capacitance depending on the weight polarity, 
and the capacitance voltage is considered a membrane potential. 
The output spikes of the previous layer control the switch of the current source, 
accumulating the spike as membrane potential, and the magnitude of the current source is determined by the weights of the SNN. 
When the membrane potential reaches a threshold voltage,
the neuron generates a spike and the membrane potential decreases by the threshold voltage.
The inherent problems of this implementation are as follows. 
$V_{DS}$ must be larger than $V_{OD}$ for the current source to be in the saturation region, and even if it is in the saturation region, the current value is changed by $V_{DS}$ due to channel length modulation as shown in Fig.\ref{fig:vdtd}(b). 
Acquiring a voltage headroom by increasing supply voltage can mitigate the problem, 
but this leads to higher power consumption, which undermines the benefits of SNN. 
Also, exploiting a cascode current source to relax channel length modulation could be a possible solution, 
but it decreases voltage headroom and reduces the dynamic range of membrane potential. 
Therefore, for voltage-domain neurons where nonlinearity inevitably occurs, it is necessary to investigate how this characteristic affects the SNN performance.

\begin{figure*}
\begin{subfigure}[b]{0.45\linewidth}
    \centering
    \includegraphics[scale=0.7]{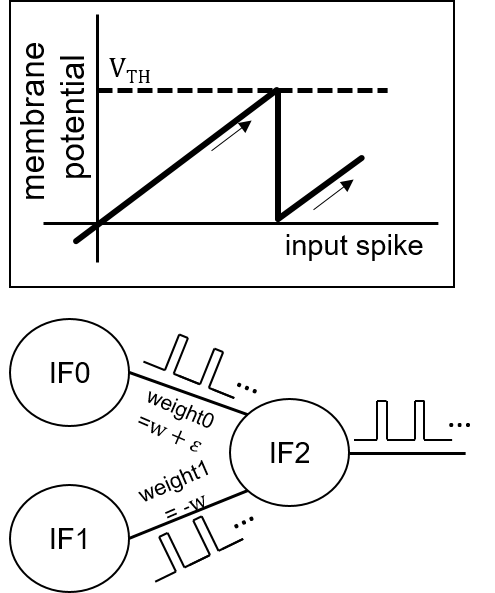}
    \caption{Spike computation with ideal neurons}
    \label{fig:spike_comp_ideal}
\end{subfigure}
\hfill
\begin{subfigure}[b]{0.45\linewidth}
    \centering
    \includegraphics[scale=0.7]{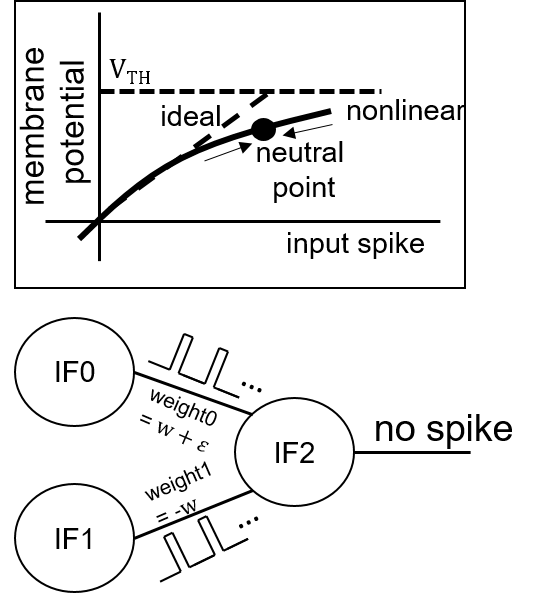}
    \caption{Spike computation with nonlinear neurons}
    \label{fig:spike_comp_nonlin}
\end{subfigure}%
\caption{Effect of nonlinear neurons on membrane potential dynamics.}
\label{fig:spike_comp}
\end{figure*}

\subsection{Impact of nonlinearity on SNN}
\label{sec:nonlinearity}

In order to understand the effect of nonlinearity on SNN, 
consider a simple network consisting of three IF-neurons as shown in Fig.~\ref{fig:spike_comp}. 
The outputs of two neurons (IF0,IF1) are applied to the next neuron (IF2), 
and the outputs of IF0 and IF1 generate spikes at the same rate.
Additionally, let us assume that weight1 is negative ($-w$) and weight0 is positive ($w+\epsilon$) as described in Fig.~\ref{fig:spike_comp}. 
If the IF-neuron is ideal (Fig.~\ref{fig:spike_comp}(a)), 
the membrane potential will increase at a constant rate 
because positive spikes are slightly stronger than negative spikes. 
It increases until the membrane potential exceeds the threshold,
and an output spike will be generated. 
Then, the membrane potential will be subtracted by the threshold voltage,
and it continues to generate spikes at a constant rate. 
 
On the other hand, if IF-neuron is nonlinear as shown in Fig.~\ref{fig:spike_comp}(b), 
the membrane potential will gradually increase as in the ideal case when the potential is near zero. 
However, even if the magnitude of weight1 is larger than that of weight0,
at a point where the positive spike and the negative spike have the same strength because of the nonlinear characteristic, 
the membrane potential will no longer increase. 
This point is expressed as "neutral point" as shown in Fig.\ref{fig:spike_comp_nonlin}. 
In this case, no spike is generated ever. 
To get rid of this nonlinearity, Section~\ref{sec:proposed} proposes a time-based neuron that fundamentally removes the voltage-domain nonlinear behavior.
\section{Proposed Spiking Neural Network}
\label{sec:proposed}

\subsection{Time-based neuron for SNN} 
SNN neurons should be able to accumulate input spikes. 
In voltage-domain neurons, the accumulation function is implemented by flowing currents into a capacitor and converting it into a voltage. 
On the other hand, time-based neurons take advantage of voltage-controlled oscillators (VCOs) and embed the membrane potential in the time, or phase, domain.
As shown in Fig.~\ref{fig:vdtd}(c), when spikes are received as inputs, VCO frequency is changed and 
accumulation of spikes can be obtained from the VCO phase.
When the phase shift reaches a certain threshold, which can be detected by a phase detector, 
the time-based neuron generates an output spike.

\subsection{Linearity of time-based neuron}
As shown in Fig.\ref{fig:vdtd}, current sources are used in the time-based neuron, which is not ideal either. 
Nevertheless, it does not cause nonlinearity in the time-based neuron 
because the drain voltage of the current source is not accumulated with spike inputs 
unlike the drain voltage in the voltage-domain neuron. 
In the time-based neuron, at the moment when the spike comes in, 
the drain voltage will rise a little, 
but when the spike disappears, the voltage will fall back to the original value. 
Therefore, the current source can flow the same amount of current regardless of the accumulated input spike. 
In addition, there is no limitation for membrane potential because phase is able to rotate indefinitely, whereas the range of membrane potential in the voltage-domain neuron is bounded by the supply voltage. 
This characteristic is also advantageous to increase the resolution of membrane potential.

\begin{figure}
    \centering
    \includegraphics[scale=0.4]{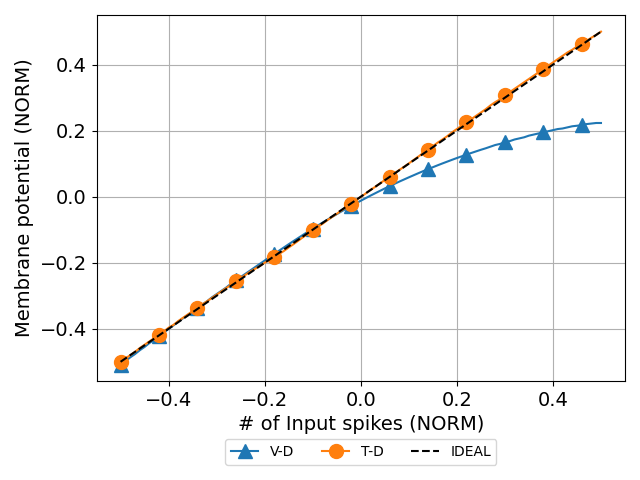}
    \caption{Simulated membrane potential via input spikes of ideal, voltage-domain, time-domain neuron.}
    \label{fig:inl}
     \vspace{-0.5em}
\end{figure}

\section{Experiments}
\label{sec:experiments}

\subsection{Circuit design and simulation}
In order to model realistic linearity characteristics, 
both voltage-domain neuron and time-domain neuron are designed in a 28\,nm process and simulated. 
The voltage-based neuron is simulated while applying spikes of constant ratio to the input, and the output voltage of the capacitor is measured. 
The time-based neuron is simulated by applying the same spikes and measuring the output phase shift. 
As a result, membrane potential via input spikes is obtained as shown in Fig.~\ref{fig:inl}. 
Note that ranges of membrane potential and the number of input spikes are normalized to -1 to 1.
As the number of input spikes to the voltage-based neuron increase, 
the amount of change in the membrane potential per spike becomes smaller due to channel length modulation.
On the other hand, in the case of time-based neuron, the amount of change in the membrane potential per spike keeps constant in the entire range just like the ideal case.

\subsection{System-level simulation with nonlinearity modeling}
\begin{figure}
    \centering
    \includegraphics[scale=0.38]{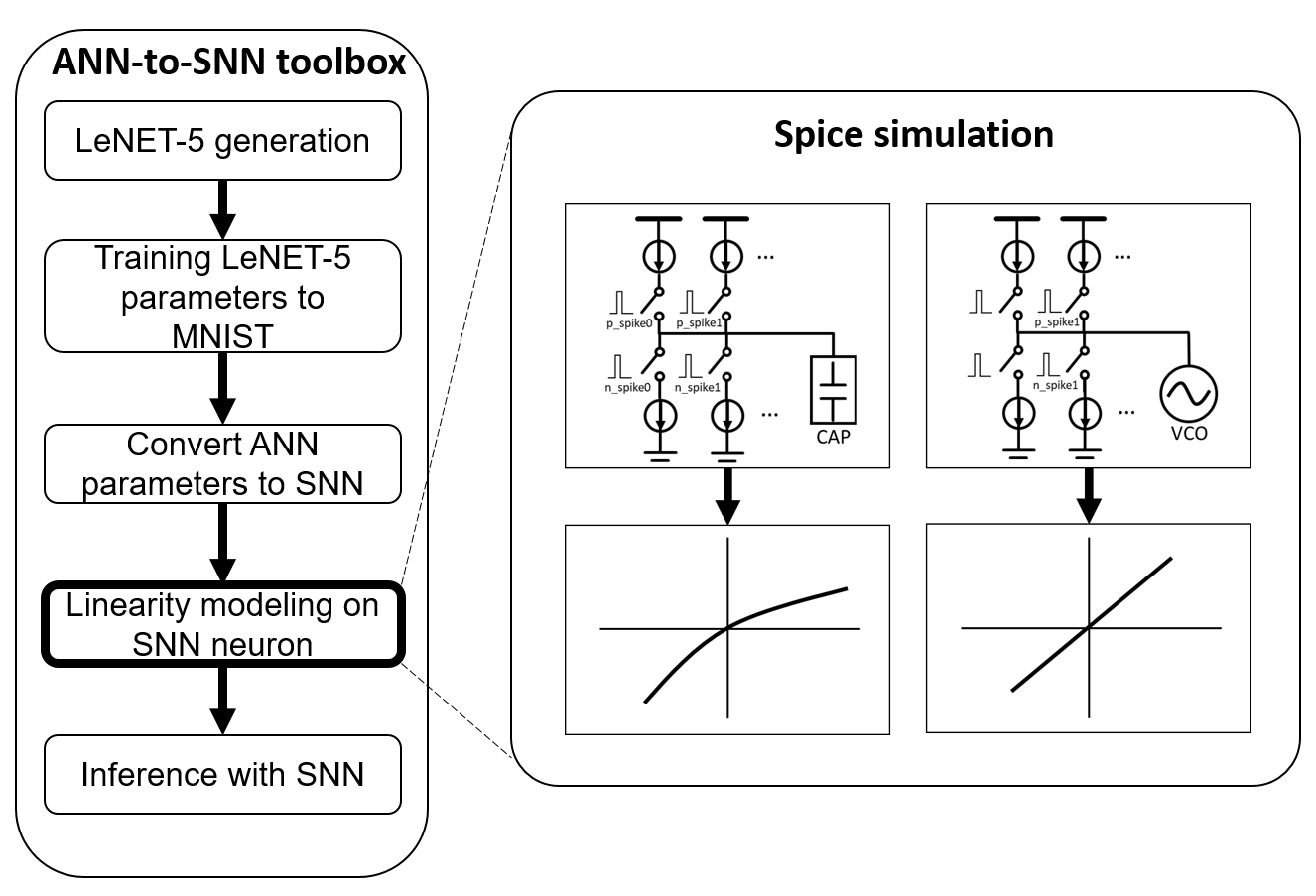}
    \caption{SNN simulation setup with linearity modeling.}
    \label{fig:sim_setup}
\end{figure}

\begin{figure}
    \centering
    \includegraphics[scale=0.4]{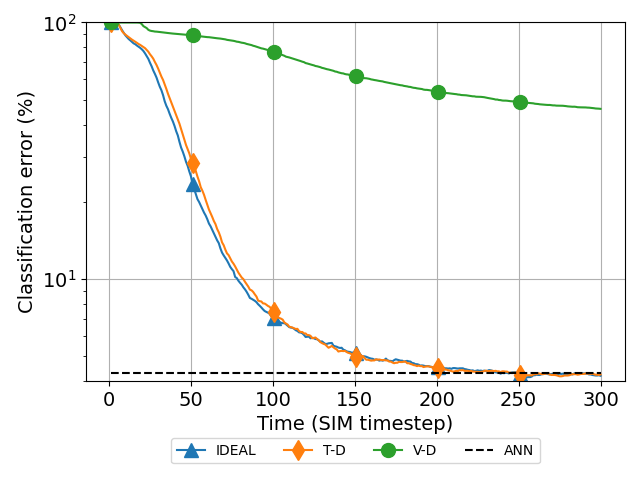}
    \caption{LeNet5 MNIST classfication with linearity modeling}
    \label{fig:le}
\end{figure}

To find out how the neuron's nonlinearity affects SNN, 
system-level simulation is performed to classify MNIST dataset with both voltage-domain and time-domain neuron model. 
For the experiments, we customized and exploited the ANN-to-SNN conversion tool, which is an open source released by \cite{Ref:3}. 
The simulation setup is configured as described in \ref{fig:sim_setup}.
LeNet-5~\cite{lecun1998gradient} is chosen for classification, 
and after training the model with MNIST, 
the trained parameters are extracted and converted into SNN layers.
Then, MNIST classification with neuron models extracted from circuit simulation is carried out. 
Three different models---voltage-based neuron, time-based neuron, and ideal neuron---are applied to the simulation and compared as shown in Fig.~\ref{fig:le}
The x-axis represents the simulation timestep and 
every simulation timestep, it computes membrane potential and determines whether a spike is fired. 
The y-axis is the classification error rate. It takes some time for error rate to reach its stable point because spikes sequence in SNN is a stochastic process and a few samples cannot represent their final value. 
After the error rates have been stabilized, 
the time-based neuron has an error rate (4.25\,\%) almost similar to that of ideal neuron (4.2\,\%). 
On the other hand, the accuracy of the voltage-based neuron is degraded.

\begin{figure}
\begin{subfigure}[b]{0.5\textwidth}
    \centering
    \includegraphics[scale=0.4]{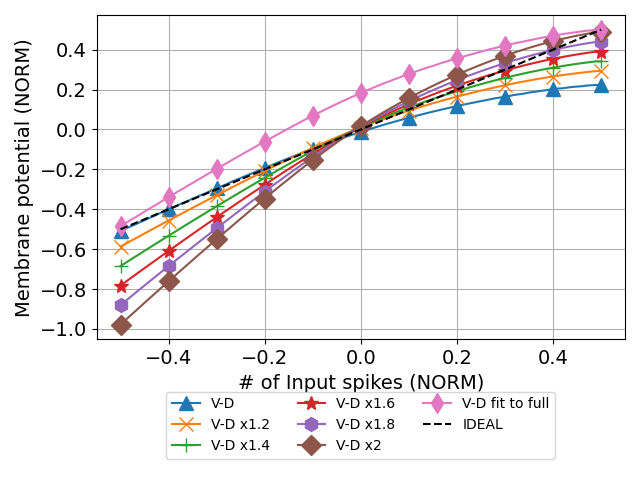}
    \caption{Re-scaled nonlinearity model of voltage-domain neuron}
    \label{fig:INL_curr}
\end{subfigure}
\hfill
\begin{subfigure}[b]{0.5\textwidth}
    \centering
    \includegraphics[scale=0.4]{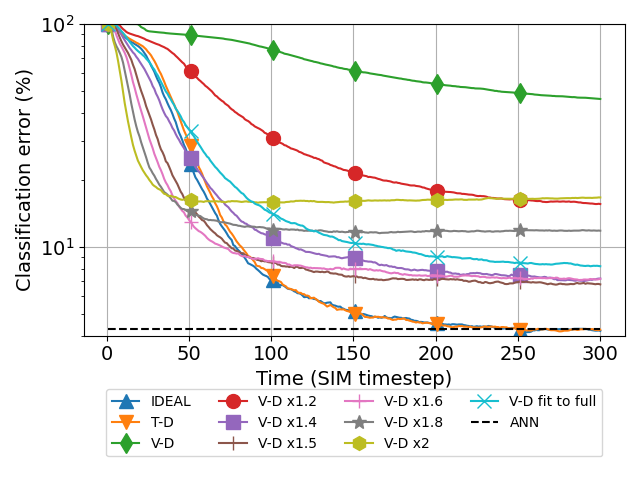}
    \caption{LeNet5 MNIST classfication with re-scaled model}
    \label{fig:le_curr}
\end{subfigure}
\hfill
\caption{Re-scaled voltage-domain neuron model and MNIST classfication.}
\label{fig:inl_le_curr}
\end{figure}

Since the performance degradation of the voltage-based neuron model may have occurred simply due to failing to cover the full range of the ideal model as shown in Fig.~\ref{fig:inl}, 
several attempts to re-scale the nonlinear transfer curve of the voltage-based neuron model are conducted.
Fig.~\ref{fig:INL_curr} shows the nonlinear models re-scaled in various ways. 
Re-scaling with constant factors (1.2, 1.4, 1.5, 1.6, 1.8, 2) and fitting to the maximum range of the ideal curve are performed.
Despite various re-scaling, all results have less performance than time-domain neuron as shown in Fig.~\ref{fig:le_curr}.
As mentioned in Section~\ref{sec:nonlinearity}, nonlinearity can only be compensated by adjusting each coefficient individually, so it cannot achieve the performance corresponding to the ideal neuron or voltage-based neuron with any simple re-scaling.
Through simulation results above, it is shown that the nonlinearity of the voltage-based neuron degrades SNN performance 
while the time-based neuron has high accuracy close to the ideal neuron.

\section{Conclusion}
\label{sec:conclusion}

In this paper, the nonlinear behavior of existing voltage-based neurons is characterized in a 28\,nm process, and we demonstrate that these nonlinear characteristics can cause significant SNN performance degradation. 
In order to overcome the analog neuron design challenges posed by deep submicron technologies, 
a time-based neuron is proposed to guarantee linear characteristics. 
Tested on the MNIST dataset, the proposed neuron achieves 4.25\,\% classification error differing less than 0.1\,\% from that of the ideal neuron.

\section*{Acknowledgment}

The EDA Tool was supported by the IC Design Education Center.

\bibliographystyle{IEEEtran}
\bibliography{Ref}

\begin{thebibliography}{10}
\providecommand{\url}[1]{#1}
\csname url@samestyle\endcsname
\providecommand{\newblock}{\relax}
\providecommand{\bibinfo}[2]{#2}
\providecommand{\BIBentrySTDinterwordspacing}{\spaceskip=0pt\relax}
\providecommand{\BIBentryALTinterwordstretchfactor}{4}
\providecommand{\BIBentryALTinterwordspacing}{\spaceskip=\fontdimen2\font plus
\BIBentryALTinterwordstretchfactor\fontdimen3\font minus
  \fontdimen4\font\relax}
\providecommand{\BIBforeignlanguage}[2]{{%
\expandafter\ifx\csname l@#1\endcsname\relax
\typeout{** WARNING: IEEEtran.bst: No hyphenation pattern has been}%
\typeout{** loaded for the language `#1'. Using the pattern for}%
\typeout{** the default language instead.}%
\else
\language=\csname l@#1\endcsname
\fi
#2}}
\providecommand{\BIBdecl}{\relax}
\BIBdecl

\bibitem{horowitz20141}
M.~Horowitz, ``Computing's energy problem (and what we can do about it),'' in
  \emph{IEEE International Solid-State Circuits Conference Digest of Technical
  Papers}.\hskip 1em plus 0.5em minus 0.4em\relax IEEE, 2014, pp. 10--14.

\bibitem{neftci2019surrogate}
E.~O. Neftci, H.~Mostafa, and F.~Zenke, ``Surrogate gradient learning in
  spiking neural networks: Bringing the power of gradient-based optimization to
  spiking neural networks,'' \emph{IEEE Signal Processing Magazine}, vol.~36,
  no.~6, pp. 51--63, 2019.

\bibitem{mostafa2017supervised}
H.~Mostafa, ``Supervised learning based on temporal coding in spiking neural
  networks,'' \emph{IEEE transactions on neural networks and learning systems},
  vol.~29, no.~7, pp. 3227--3235, 2017.

\bibitem{shrestha2018slayer}
S.~B. Shrestha and G.~Orchard, ``Slayer: Spike layer error reassignment in
  time,'' in \emph{NeurIPS}, 2018.

\bibitem{wu2018spatio}
Y.~Wu, L.~Deng, G.~Li, J.~Zhu, and L.~Shi, ``Spatio-temporal backpropagation
  for training high-performance spiking neural networks,'' \emph{Frontiers in
  neuroscience}, vol.~12, p. 331, 2018.

\bibitem{Ref:4}
J.~H. Lee, T.~Delbruck, and M.~Pfeiffer, ``Training deep spiking neural
  networks using backpropagation,'' \emph{Frontiers in Neuroscience}, vol.~10,
  2016.

\bibitem{Ref:3}
B.~Rueckauer, I.-A. Lungu, Y.~Hu, M.~Pfeiffer, and S.-C. Liu, ``Conversion of
  continuous-valued deep networks to efficient event-driven networks for image
  classification,'' \emph{Frontiers in neuroscience}, vol.~11, p. 682, 2017.

\bibitem{hwang2020impact}
S.~Hwang, J.~Chang, M.-H. Oh, J.-H. Lee, and B.-G. Park, ``Impact of the
  sub-resting membrane potential on accurate inference in spiking neural
  networks,'' \emph{Scientific reports}, vol.~10, no.~1, pp. 1--10, 2020.

\bibitem{Ref:1}
J.~Park, M.-W. Kwon, H.~Kim, S.~Hwang, J.-J. Lee, and B.-G. Park, ``Compact
  neuromorphic system with four-terminal si-based synaptic devices for spiking
  neural networks,'' \emph{IEEE Transactions on Electron Devices}, vol.~64,
  no.~5, pp. 2438--2444, 2017.

\bibitem{huh2017gradient}
D.~Huh and T.~J. Sejnowski, ``Gradient descent for spiking neural networks,''
  \emph{arXiv preprint arXiv:1706.04698}, 2017.

\bibitem{Ref:5}
E.~Stromatias, M.~Soto, T.~Serrano-Gotarredona, and B.~Linares-Barranco, ``An
  event-driven classifier for spiking neural networks fed with synthetic or
  dynamic vision sensor data,'' \emph{Frontiers in Neuroscience}, vol.~11,
  2017.

\bibitem{Ref:6}
Y.~Cao, Y.~Chen, and D.~Khosla, ``Spiking deep convolutional neural networks
  for energy-efficient object recognition,'' \emph{International Journal of
  Computer Vision}, 2015.

\bibitem{Ref:7}
P.~U. Diehl, D.~Neil, J.~Binas, M.~Cook, S.-C. Liu, and M.~Pfeiffer,
  ``Fast-classifying, high-accuracy spiking deep networks through weight and
  threshold balancing,'' \emph{International Joint Conference on Neural
  Networks}, 2015.

\bibitem{lecun1998gradient}
Y.~LeCun, L.~Bottou, Y.~Bengio, and P.~Haffner, ``Gradient-based learning
  applied to document recognition,'' \emph{Proceedings of the IEEE}, vol.~86,
  no.~11, pp. 2278--2324, 1998.

\end{thebibliography}

\end{document}